\documentclass[letterpaper]{article} 
\usepackage[]{aaai2026}  
\usepackage{times}  
\usepackage{helvet}  
\usepackage{courier}  
\usepackage[hyphens]{url}  
\usepackage{graphicx} 
\usepackage{natbib}  
\usepackage{caption} 
\usepackage{algorithm}
\usepackage{algpseudocode}      

\usepackage{amsmath}            
\usepackage{amssymb}            
\usepackage{amsfonts}           
\usepackage{multirow}           
\usepackage{booktabs}           
\usepackage{makecell}           
\usepackage{tabularx}           
\usepackage{arydshln}           
\usepackage{graphicx}           
\usepackage{float}              
\usepackage{algorithm}          
\usepackage{algpseudocode}      
\usepackage{bm}                 
\usepackage[tikz]{bclogo}       
\usepackage{caption}
\usepackage{color}              
\usepackage{enumitem}           

\newcommand{\Benchmark}{\textsc{CL$^2$GEC}}

\definecolor{brickred}{HTML}{b92622}
\definecolor{midnightblue}{HTML}{005c7f}
\definecolor{salmon}{HTML}{f1958d}
\definecolor{burntorange}{HTML}{f19249}
\definecolor{junglegreen}{HTML}{4dae9d}
\definecolor{forestgreen}{HTML}{499c5e}
\definecolor{pinegreen}{HTML}{3d8a75}
\definecolor{seagreen}{HTML}{6bc1a2}
\definecolor{limegreen}{HTML}{97c65a}

\definecolor{Evidence}{RGB}{255,0,0}
\definecolor{Linguistic}{RGB}{84,130,53}
\definecolor{Cause}{RGB}{0,176,240}
\definecolor{Revision}{RGB}{31,78,121}

\usepackage{newfloat}
\usepackage{listings}
\DeclareCaptionStyle{ruled}{labelfont=normalfont,labelsep=colon,strut=off} 
\floatstyle{ruled}
\newfloat{listing}{tb}{lst}{}
\floatname{listing}{Listing}
\title{\Benchmark{}: A Multi-Discipline Benchmark for Continual Learning in \\Chinese Literature Grammatical Error Correction}

\author{
\textbf{Shang Qin}$^{1}$\thanks{indicates equal contribution.},
\textbf{Jingheng Ye}$^{1*}$,
\textbf{Yinghui Li}$^{1}$,
\textbf{Hai-Tao Zheng}$^{1,2}$\thanks{Corresponding author: Hai-Tao Zheng. (E-mail: zheng.haitao@sz.tsinghua.edu.cn)}, \\
\textbf{Qi Li}$^{3}$, 
\textbf{Jinxiao Shan}$^{3}$,
\textbf{Zhixing Li}$^{4}$,  
\textbf{Hong-Gee Kim $^{5}$}\\  
$^{1}$Tsinghua Shenzhen International Graduate School, Tsinghua University  \\
$^{2}$Peng Cheng Laboratory, 
$^{3}$China Merchants Group,
$^{4}$Zhipu AI \\
$^{5}$Seoul National University \\
}

\begin{document}

\maketitle

\begin{abstract}
The growing demand for automated writing assistance in diverse academic domains highlights the need for robust Chinese Grammatical Error Correction (CGEC) systems that can adapt across disciplines. However, existing CGEC research largely lacks dedicated benchmarks for multi-disciplinary academic writing, overlooking continual learning (CL) as a promising solution to handle domain-specific linguistic variation and prevent catastrophic forgetting. To fill this crucial gap, we introduce \textbf{\Benchmark{}}, the first \underline{C}ontinual \underline{L}earning benchmark for \underline{C}hinese \underline{L}iterature \underline{G}rammatical \underline{E}rror \underline{C}orrection, designed to evaluate adaptive CGEC across multiple academic fields. Our benchmark includes 10,000 human-annotated sentences spanning 10 disciplines, each exhibiting distinct linguistic styles and error patterns.  \Benchmark{} focuses on evaluating grammatical error correction in a continual learning setting, simulating sequential exposure to diverse academic disciplines to reflect real-world editorial dynamics. We evaluate large language models under sequential tuning, parameter-efficient adaptation, and four representative CL algorithms, using both standard GEC metrics and continual learning metrics adapted to task-level variation. Experimental results reveal that regularization-based methods mitigate forgetting more effectively than replay-based or naive sequential approaches. Our benchmark provides a rigorous foundation for future research in adaptive grammatical error correction across diverse academic domains.
\end{abstract}

\section{Introduction}\label{sec:introduction}

Chinese Grammatical Error Correction (CGEC) has evolved rapidly alongside the surge of large language models (LLMs)~\citep{ye2025corrections, li2024benchmarking, yu2024seqgpt, qingsong2025raise, li2025one, li2025refine, kuang2025express, li2025mdit, chen2025dast} and intelligent writing assistants~\cite{DBLP:conf/acl/LiZLLLSWLCZ22, DBLP:conf/emnlp/LiMZLLHLLC022, li2023ineffectiveness,li2023towards, qiu2025chinese,li2024rethinking, ye-etal-2023-cleme, zhang2023contextual, zhang2025loss}. Most existing CGEC benchmarks, however, are (1) learner or general domain oriented~\cite{zhang2022mucgec,DBLP:conf/emnlp/MaLSZHZLLLCZS22}, and (2) evaluated in a static setting~\cite{xu-etal-2022-fcgec,ye-etal-2023-cleme,ye2024cleme2}. As a result, they offer limited insight into how CGEC models behave in high-stakes professional domains, especially in scientific manuscripts where style, terminology, and error distribution vary markedly across disciplines.

\begin{figure}[tb!]
\centering
\includegraphics[width=0.9\linewidth]{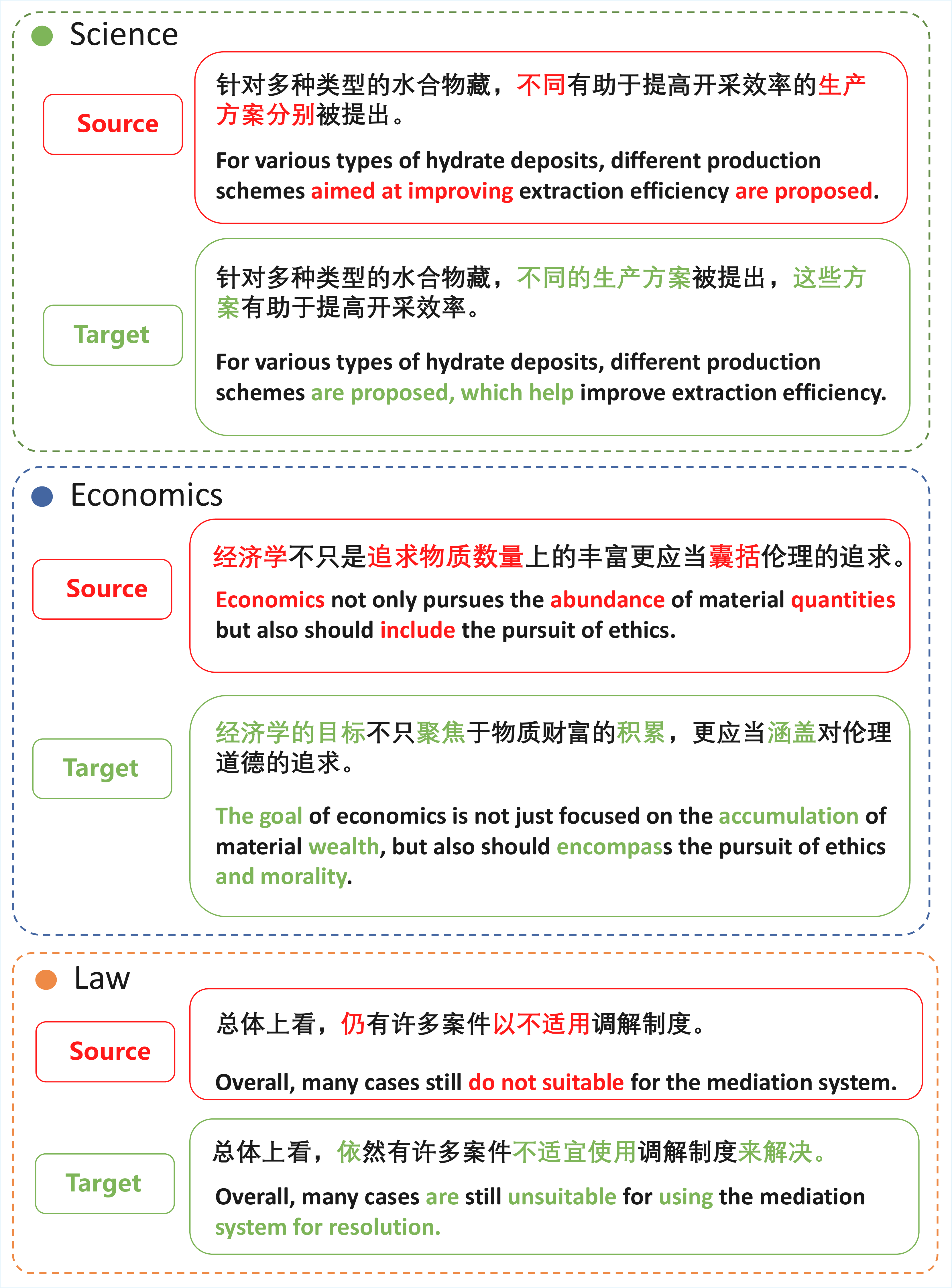}
\caption{
The correction examples from three academic disciplines (Science, Economics, Law) in \Benchmark{}.}
\label{fig:intro}
\end{figure}

We argue that real-world scientific writing introduces an under-explored challenge for CGEC: \textit{continual domain adaptation}~\cite{wu2024continual,guan-etal-2025-multi}. In practice, CGEC systems is expected to ingest papers from, e.g., Physics this month and Humanities next month, continually refining its internal knowledge without access to all past data. The threat of catastrophic forgetting~\cite{zhai2023investigating,li-etal-2024-revisiting,huang-etal-2024-mitigating}, widely studied in vision~\cite{shmelkov2017incremental} and NLP tasks~\cite{shao-feng-2022-overcoming}, has received almost no attention in CGEC, leaving an open question: \emph{Can modern LLMs retain grammatical knowledge while sequentially adapting to new scientific disciplines?}

Addressing this question is crucial for three reasons. First, academia encompasses hundreds of sub-fields whose linguistic conventions differ in syntax and terminology, significantly challenging current LLMs. Second, annotation budgets are usually fragmented by discipline, making one-shot and full-data retraining impractical. Third, reliable cross-domain grammars underpin downstream tasks such as automatic reviewing~\cite{pang2025understanding}, plagiarism detection~\cite{quidwai2023beyond}, and literature summarization~\cite{li2024chatcite}.


Therefore, to systematically study CGEC under the context of continual learning (CL), we present \textbf{\Benchmark{}}, the first \underline{C}ontinual \underline{L}earning benchmark for \underline{C}hinese \underline{L}iterature \underline{G}rammatical \underline{E}rror \underline{C}orrection. The \Benchmark{} benchmark contains 10,000 sentences evenly sampled from 10 academic disciplines, each sentence paired with up to three independent human references. The corpus was curated from China National Knowledge Infrastructure (CNKI)\footnote{\url{https://www.cnki.net/}}, cleaned for copyright, and double-checked by professional editors to reflect authentic error patterns. We release both a canonical split (train/dev/test) and a sequence of 10 task partitions that simulate chronological arrival of disciplines, enabling controlled CL evaluation. \Benchmark{} aims to set a new standard for evaluating and advancing lifelong grammatical error correction in the era of domain-diverse scientific communication.

Our proposed \Benchmark{} allows researchers to probe a spectrum of model abilities, mainly including (1) in-domain grammatical accuracy, (2) cross-discipline transfer, and (3) resistance to catastrophic forgetting. Consequently, the benchmark fills a vital gap between generic CGEC datasets and real-world academic editing.

Empirically, we benchmark several representative continual learning strategies, including naive sequential fine-tuning, LoRA-based adaptation~\citep{hu2021loralowrankadaptationlarge}, and four CL algorithms (EWC~\citep{Kirkpatrick_Pascanu_Rabinowitz_Veness_Desjardins_Rusu_Milan_Quan_Ramalho_Grabska-Barwinska_et_al._2017}, GEM~\citep{Lopez-Paz_Ranzato_2017}, LwF~\citep{Li_Hoiem_2016}, OGD~\citep{Farajtabar_Azizan_Mott_Li_2019}). Our comprehensive experiments reveal that while these methods significantly mitigate catastrophic forgetting compared to naive sequential approaches, the optimal strategies vary. We observe a nuanced impact of task ordering on knowledge retention and transfer, including unexpected interference between semantically related disciplines and a trade-off between precision and recall depending on the task sequence. 

\textbf{Our contributions are as follows:}
\begin{itemize}
  \item We introduce \Benchmark{}, the first large-scale and multi-discipline benchmark tailored to Chinese literature grammatical error correction in the context of continual learning.
  \item We devise task-specific CL metrics (Average Performance, Backward/Forward Transfer, Forgetting Rate) adapted to CGEC, and provide a reproducible evaluation suite.
  \item We conduct extensive LLM experiments, revealing critical limitations of existing CL methods and establishing solid baselines for future work.
\end{itemize}

\section{Related Work}\label{sec:related_work}

\subsection{Chinese Grammatical Error Correction}

Chinese grammatical error correction (CGEC) has developed from early sequence-to-sequence (Seq2Seq) models~\citep{ye2022focus, DBLP:journals/csur/DongLGCLSY23, DBLP:journals/patterns/LiuLTLZ22, zhao2019cged, kaneko-etal-2020-encoder, zhang2022mucgec, DBLP:conf/emnlp/YeLL023, ye2025excgec, li2025correct}, which model correction as a generation task. These approaches benefited from pretraining and syntactic priors but were mainly applied to general learner corpora.

With the advent of large language models (LLMs)~\citep{huang2023lateval, kuang2025natural, DBLP:journals/corr/abs-2402-11100,DBLP:journals/corr/abs-2402-19248, du2024llms}, recent work has explored their capabilities for CGEC~\citep{wang-etal-2024-towards-realistic, xiao-etal-2024-chinese, DBLP:conf/emnlp/HuangYZLLZZ23, ye2023system}. Closed- and open-source LLMs have been evaluated through in-context learning and instruction tuning, showing improved fluency and generalization across error types. ScholarGEC~\citep{li2025ScholarGEC} further investigates controllability in academic writing by combining error detection and correction within a multi-task framework.

While prior work focuses on enhancing model performance under static settings, relatively little attention has been paid to domain transfer or continual adaptation. Our CSLGEC benchmark addresses this gap by providing a multi-discipline scientific dataset and evaluating models under domain-incremental settings, enabling more realistic and systematic assessment of generalization in CGEC.




\subsection{Continual Learning for NLP}

Continual learning (CL) aims to enable models to learn from sequential data without forgetting previously acquired knowledge~\citep{delange2021continual, xing2024mitigating}. Existing CL methodologies can be broadly grouped into three primary categories: regularization-based, replay-based, and architectural-based approaches.

\textbf{Regularization-based} methods constrain model parameter updates to mitigate forgetting. They typically introduce penalty terms to the loss function to protect the values of critical weights from earlier tasks. This category encompasses several prominent methods evaluated in our study. For instance, Orthogonal Gradient Descent (OGD)~\citep{Farajtabar_Azizan_Mott_Li_2019} enforces that a task's gradient updates are orthogonal to those of prior tasks, thereby preventing interference. Elastic Weight Consolidation (EWC)~\citep{Kirkpatrick_Pascanu_Rabinowitz_Veness_Desjardins_Rusu_Milan_Quan_Ramalho_Grabska-Barwinska_et_al._2017} identifies and safeguards the most salient parameters from past tasks using a quadratic penalty. Gradient Episodic Memory (GEM)~\citep{Lopez-Paz_Ranzato_2017} leverages a small episodic memory buffer to ensure that the current task's loss is not increased on previous tasks' examples, effectively avoiding backward loss. Another notable method, Learning without Forgetting (LwF)~\citep{Li_Hoiem_2016}, employs knowledge distillation from a frozen copy of the previous model to guide learning on the new task.

\textbf{Replay-based} methods prevent forgetting by re-exposing the model to data from old tasks during training on new ones. The foundational strategy in this approach is Experience Replay (Replay), which uses a buffer to store and replay samples from past tasks as a strong baseline. More sophisticated techniques involve generative replay, which uses generative models to synthesize data, or methods that capture statistical information of old tasks instead of replaying the original data. SEEKR~\citep{he-etal-2024-seekr}, a replay-based distillation approach that preserves important attention mechanisms by introducing a unique head importance measure that accounts for forgettability.

Finally, \textbf{Architectural-based} approaches tackle forgetting by altering the model's architecture to integrate new information. A prime example is Progressive Prompts~\citep{Razdaibiedina_Mao_Hou_Khabsa_Lewis_Almahairi_2023}, which learns distinct prompts for each task and then sequentially chains them to form a composite prompt.

However, grammatical error correction poses distinct challenges, requiring outputs that are accurate, fluent, and domain-sensitive. Despite its importance, CL has not been systematically studied in the context of CGEC. Our work fills this gap by introducing the first continual CGEC benchmark, enabling the evaluation of models under sequential exposure to diverse academic disciplines.

\section{\Benchmark{} Benchmark}
\label{sec:benchmark}

\subsection{Problem Definition}
\paragraph{Grammatical Error Correction}
Grammatical Error Correction (GEC) aims to transform an ungrammatical sentence $X = \{x_1, x_2, \dots, x_T\}$ into its grammatically correct counterpart $Y = \{y_1, y_2, \dots, y_{T'}\}$ while preserving the original semantics. Typically formulated as a sequence-to-sequence task, GEC models are trained to minimize the negative log-likelihood of the corrected output:
\begin{equation}
    \mathcal{L}_{\operatorname{GEC}} = -\sum_{t=1}^{T'} \log P(y_t \mid Y_{<t}, X).
\end{equation}

\paragraph{Continual Learning}
Continual learning (CL) addresses the challenge of learning from a stream of tasks $\{\mathcal{D}_1, \dots, \mathcal{D}_N\}$ without access to previous task data. In the supervised setting, each task $\mathcal{D}_t = \{(x_i^t, y_i^t)\}_{i=1}^{n_t}$ is presented sequentially, and the model is trained to accumulate knowledge over time while avoiding catastrophic forgetting. Let $f_\Theta$ be a model with parameters $\Theta$. The goal of CL is to optimize performance across all tasks:
\begin{equation}
    \max_{\Theta} \sum_{t=1}^{N} \sum_{(x,y) \in \mathcal{D}_t} \log P_\Theta(y \mid x).
\end{equation}
Evaluation in CL involves metrics such as \textit{Average Performance} and \textit{Backward Transfer}, which measure the model's ability to retain and transfer knowledge across tasks.

\paragraph{The CSLGEC Task}  
We define \Benchmark{} as a domain-specific Grammatical Error Correction (GEC) benchmark formulated under the continual learning (CL) paradigm. The dataset is composed of grammatically erroneous academic sentences collected from 10 distinct disciplines (e.g., law, medicine, philosophy), with each domain corresponding to a sequential task:
\begin{equation}
    \mathcal{D}_t = \{(X_i^t, Y_i^t)\}_{i=1}^{n_t}.
\end{equation}
Each pair consists of an erroneous sentence \(X_i^t\) and its corrected version \(Y_i^t\).
Unlike typical GEC tasks trained on all data simultaneously, CSLGEC is formulated as a continual learning benchmark where the model learns sequentially from each domain and only replays a small subset of previous data. This setup simulates real-world constraints such as limited storage and privacy, requiring the model to maintain grammatical correction performance across diverse domains without catastrophic forgetting.

\subsection{Benchmark Construction}

\paragraph{Data Collection.}
We crawl full-text PDFs from the China National Knowledge Infrastructure (CNKI)\footnote{\url{https://www.cnki.net/}}, the largest Chinese academic repository. To capture broad domain diversity, we target 10 first-level disciplines: \textit{Law, Management, Education, Economics, Science, History, Agriculture, Literature, Art}, and \textit{Philosophy}. The dataset is designed to be highly diverse and multi-disciplinary, ensuring comprehensive coverage of academic writing. These 10 first-level disciplines are further subdivided into a total of 100 second-level disciplines, providing a granular representation of academic writing. For instance, the Agriculture discipline includes sub-disciplines such as Agricultural Resources and Environment (164 instances). The detailed breakdown of all disciplines and their instance counts can be found in the Table~\ref{tab:discipline_breakdown_full}. For each discipline, we randomly sample 1,000 sentences, yielding a balanced corpus of 10,000 instances. This one-to-one ratio eliminates domain-size bias and guarantees that subsequent continual-learning curricula are not dominated by any single field. To ensure a standardized evaluation, we provide a canonical split for each discipline: 800 training examples, 100 development examples, and 100 test examples.

\begin{table}[htbp]
\renewcommand{\arraystretch}{1.2}
\renewcommand{\tabcolsep}{6pt}
\centering
\small
\begin{tabular}{lc}
    \toprule
    \textbf{First-Level Discipline} & \textbf{No. of Secondary Disciplines} \\
    \midrule
    Agriculture & 20 \\
    History     & 14 \\
    Philosophy  & 12 \\
    Education   & 5  \\
    Literature  & 5  \\
    Law         & 17 \\
    Science     & 23 \\
    Management  & 17 \\
    Economics   & 18 \\
    Art         & 3  \\
    \bottomrule
\end{tabular}
\caption{Number of Secondary Disciplines per First-Level Category.}
\label{tab:discipline_breakdown_full}
\end{table}


\paragraph{Data Cleaning.}
Because CNKI only provides PDF files, a dedicated preprocessing pipeline is required. The overall cleaning procedure is executed by a trained annotation team and guarantees that every remaining sentence is grammatically self-contained and suitable for correction.

\begin{enumerate}[leftmargin=*]
\item \textbf{PDF $\to$ JSON conversion}. We convert each PDF into a structured JSON file that preserves sentence boundaries, section tags, and positional metadata.
This machine-readable format facilitates downstream filtering and reproducibility.

\item \textbf{Section filtering}. Only the \textit{abstract} and \textit{main body} are retained. Other sections like references, acknowledgements are discarded. These sections contain the bulk of scientific exposition and therefore the majority of grammar-related errors relevant to writing assistance.

\item \textbf{Sentence segmentation}. The retained text is split into sentence-level units using \textit{LTP}~\cite{che-etal-2021-n}, enabling sentence-level GEC evaluation.

\item \textbf{Noise removal}. Inline citations, sub-titles, mathematical equations, tables, figures, and their captions are stripped. Eliminating non-linguistic tokens avoids misleading the error-detection models and prevents annotators from spending time on irrelevant content.

\item \textbf{Anonymisation}. All personal, institutional, and document identifiers are redacted to comply with privacy regulations and facilitate open release.

\end{enumerate}

\paragraph{Data Annotation.}
Given the low error density in scholarly writing, fully manual annotation would be prohibitively expensive. We therefore adopt a human-in-the-loop strategy that combines automatic grammatical error detection, LLM pre-correction, human annotation, and expert validation.

\begin{enumerate}[leftmargin=*]
\item \textbf{Automatic grammatical error detection}. 6 well-trained CGEC models (including GECToR~\cite{omelianchuk-etal-2020-gector} and fine-tuned Chinese-BART~\cite{shao2021cpt}) are first applied to the cleaned corpus. Only sentences flagged \emph{consistently} by \emph{all} detectors are kept. This consensus voting filters out roughly 95\% of grammatically correct lines, concentrating annotation effort on the 5\% most error-prone candidates and dramatically cutting both LLM invocation and human labor.

\item \textbf{LLM pre-correction}. The shortlisted sentences are passed to GPT-4o~\cite{hurst2024gpt}, which produces a candidate correction for each error span. These machine suggestions serve as weak references, giving annotators a standardising correction style across workers.

\item \textbf{Human annotation}. We recruit senior undergraduates or graduates with majors that \emph{match} the corresponding discipline, ensuring domain awareness. After extensive training on pilot samples, annotators correct each sentence while consulting the detector output and GPT-4o suggestions. Every instance is independently revised by at least two annotators, which both improves recall and exposes stylistic alternatives.

\item \textbf{Expert validation}. Domain experts (including the paper authors) perform 100\% manual review of the double-annotated data. They refine erroneous edits, reconcile conflicts, and may add supplementary references when multiple acceptable rewrites exist.
The outcome is a high-precision, multi-reference gold annotation set.

\end{enumerate}

This multi-stage pipeline maximises annotation quality \textit{and} cost-effectiveness: automatic error detection minimises wasted effort, LLM pre-correction accelerates human editing, dual annotation guarantees inter-annotator agreement, and expert review delivers publication-grade reliability.

\subsection{Evaluation Procedure and Metrics}
We evaluate model performance in a continual learning setting, using a combination of standard GEC and continual learning metrics.

\subsubsection{Evaluation Protocol}
Let $\{T_1, \dots, T_N\}$ denote the sequence of $N=10$ tasks, corresponding to our 10 academic disciplines. In this continual learning setting, models are trained sequentially on discipline-specific training sets. To assess the impact of task order, we evaluate our models under two distinct training sequences: \emph{an ordered sequence sorted by semantic similarity} and \emph{a randomizedly order training sequence}. This allows us to measure the model's robustness to different levels of domain shift.

After learning each task $T_i$, the model is evaluated on all tasks from $T_1$ up to $T_i$. We record the performance scores:
\begin{itemize}
    \item $R_{i,i}$: GEC performance on the current task $T_i$ immediately after training.
    \item $R_{i,j}$ ($j < i$): GEC performance on a past task $T_j$ after training on $T_i$.
    \item $R_{N,j}$: The final GEC performance on each task $T_j$ after completing the entire sequence of $N$ tasks.
\end{itemize}

\subsubsection{Continual Learning Metrics}

To quantify knowledge retention and forgetting in the continual learning setting, we use two key metrics:

\paragraph{Backward Transfer (BWT)}
\begin{equation}
\text{BWT} = \frac{1}{N - 1} \sum_{j=1}^{N-1} \left( R_{N,j} - R_{j,j} \right).
\end{equation}
Backward Transfer quantifies the average change in performance on past tasks after training on all subsequent tasks. A negative BWT value indicates catastrophic forgetting, while a non-negative value suggests the model has retained or improved its knowledge of prior tasks.

\paragraph{Average Task Performance (AvgPerf)}
\begin{equation}
\text{AvgPerf} = \frac{1}{N} \sum_{j=1}^{N} R_{N,j}.
\end{equation}
Average Task Performance measures the model’s overall GEC ability across all disciplines after the full sequential training process is complete, providing a global view of the model's final competence.

\subsubsection{Standard GEC Metrics}

We adopt the ChERRANT scorer~\cite{zhang2022mucgec} to compute standard grammatical error correction (GEC) metrics, including Precision (P), Recall (R), and F$_{0.5}$. ChERRANT extends the ERRANT framework to Chinese by using character-level alignment and edit classification, enabling more accurate evaluation of corrections in Chinese text.

For each evaluation result $R_{i,j}$ (i.e., model after learning task $i$ evaluated on task $j$), we compute the P/R/F$_{0.5}$ scores using ChERRANT. The scorer identifies and classifies character-level edits between system outputs and human references, and uses tagged edits to determine true positives, false positives, and false negatives.

To summarize performance over all tasks, we first compute the average Precision and Recall across tasks, and then compute the final F$_{0.5}$ score using:
\begin{equation}\begin{aligned}
\text{F}_{0.5} = \frac{(1+0.5^2)\times \overline{P} \times \overline{R}}{0.5^2 \times \overline{P} + \overline{R}}.
\end{aligned}\end{equation}

These metrics collectively offer a holistic view of the model's performance under continual adaptation pressure, encompassing both GEC-specific and continual learning-specific aspects.

\section{Experiments}\label{sec:experiments}
\begin{table*}[htb!]
\centering
\setlength{\tabcolsep}{4pt}
\renewcommand{\arraystretch}{1.3}

\resizebox{\textwidth}{!}{%
\begin{tabular}{ll
                *{3}{c}  
                *{3}{c}  
                *{3}{c}  
                *{3}{c}  
                *{3}{c}  
                *{3}{c}  
               }
\toprule
\multirow{2}{*}{\textbf{Model}} & \multirow{2}{*}{\textbf{Strategy}}
  & \multicolumn{3}{c}{\textbf{GEC (Rnd)}} & \multicolumn{3}{c}{\textbf{GEC (Sem)}}
  & \multicolumn{3}{c}{\textbf{AvgPerf (Rnd)}} & \multicolumn{3}{c}{\textbf{AvgPerf (Sem)}}
  & \multicolumn{3}{c}{\textbf{BWT (Rnd)}}    & \multicolumn{3}{c}{\textbf{BWT (Sem)}}    \\
\cmidrule(lr){3-5}\cmidrule(lr){6-8}
\cmidrule(lr){9-11}\cmidrule(lr){12-14}
\cmidrule(lr){15-17}\cmidrule(lr){18-20}
 & 
  & P & R & F$_{0.5}$ & P & R & F$_{0.5}$ 
  & P & R & F$_{0.5}$ & P & R & F$_{0.5}$ 
  & P & R & F$_{0.5}$ & P & R & F$_{0.5}$ \\
\midrule
\multirow{7}{*}{\makecell{\textbf{Qwen2.5}\\\textbf{7B-Instruct}}}
  & SeqFT  & 59.25 & 10.71 & 29.91 & 52.10 & 12.61 & 31.08
           & 50.92 & 11.97 & 29.57 & 46.70 & 14.84 & 31.18
           & \textbf{8.13} & -1.20 & -0.40 & 0.95 & -0.13 & 0.63 \\
\cdashline{3-20}[1pt/1pt]
  & LoRA   & 65.42 & 13.00 & 35.54 & 64.18 & 13.03 & 34.83
           & 62.80 & 11.33 & 32.02 & 61.70 & 12.92 & 33.94
           & 4.54 & 1.61 & 4.01 & 1.77 & \textbf{1.55} & \textbf{3.00} \\
\cdashline{3-20}[1pt/1pt]
  & Replay & 58.78 & 11.49 & 31.13 & 47.04 & 11.22 & 26.90
           & 50.85 & 12.58 & 29.93 & 48.00 & 13.89 & 30.89
           & 5.75 & -1.73 & -0.65 & -4.31 & -1.64 & -4.17 \\
\cdashline{3-20}[1pt/1pt]
  & EWC    & 67.34 & 13.00 & 35.52 & 64.26 & 12.97 & 34.76
           & \textbf{64.88} & 11.40 & 32.11 & 61.70 & 12.93 & 33.94
           & 4.44 & 1.60 & 4.00 & 1.94 & 1.40 & 2.77 \\
\cdashline{3-20}[1pt/1pt]
  & GEM    & 67.33 & 13.10 & \textbf{35.65} & \textbf{64.34} & 13.00 & 34.82
           & 64.86 & 11.42 & 32.17 & 61.77 & 12.93 & 33.96
           & 4.41 & \textbf{1.70} & \textbf{4.11} & \textbf{1.99} & 1.45 & 2.84 \\
\cdashline{3-20}[1pt/1pt]
  & LwF    & 62.86 & \textbf{13.22} & 34.89 & 63.73 & 12.98 & 34.75
           & 58.60 & \textbf{12.84} & \textbf{33.36} & 61.36 & 13.84 & 35.24
           & 3.10 & 0.01 & 0.69 & 0.95 & 0.50 & 0.96 \\
\cdashline{3-20}[1pt/1pt]
  & OGD    & \textbf{67.78} & 12.30 & 34.44 & 62.77 & \textbf{14.43} & \textbf{36.53}
           & 63.15 & 13.38 & 34.97 & \textbf{62.07} & \textbf{15.18} & \textbf{37.08}
           & 4.85 & -1.32 & -0.81 & -1.71 & 0.78 & 0.61 \\
\midrule
\multirow{7}{*}{\makecell{\textbf{LLaMA3}\\\textbf{8B-Instruct}}}
  & SeqFT  & 45.40 & 9.10 & 24.14 & 45.60 & 9.98 & 26.01
           & 35.15 & 10.73 & 22.52 & 36.56 & 11.03 & 24.00
           & 6.11 & -2.02 & -1.32 & 4.03 & -0.24 & 0.78 \\
\cdashline{3-20}[1pt/1pt]
  & LoRA   & 64.21 & 11.11 & 31.62 & 56.80 & 12.98 & 33.03
           & 58.93 & 11.36 & 30.98 & 56.32 & 12.85 & 32.32
           & 7.51 & -0.22 & 0.64 & -1.67 & 1.97 & 2.62 \\
\cdashline{3-20}[1pt/1pt]
  & Replay & 37.31 & 10.07 & 23.7 & 34.32 & 9.56 & 22.17
           & 30.00 & 10.39 & 20.79 & 27.90 & 10.77 & 20.46
           & 7.41 & -0.60 & \textbf{2.50} & \textbf{7.73} & -0.52 & 2.64 \\
\cdashline{3-20}[1pt/1pt]
  & EWC    & 63.82 & 11.00 & 31.34 & 57.06 & 13.11 & 33.29
           & 58.80 & 11.33 & 30.91 & 56.35 & \textbf{12.91} & \textbf{32.38}
           & 7.63 & -0.22 & 0.67 & -2.34 & 1.96 & 2.55 \\
\cdashline{3-20}[1pt/1pt]
  & GEM    & \textbf{64.46} & 11.12 & 31.71 & 57.94 & 13.16 & 33.53
           & \textbf{58.97} & 11.35 & 30.99 & 56.28 & 12.82 & 32.26
           & \textbf{8.13} & \textbf{-0.12} & 0.96 & -0.67 & 2.19 & 3.16 \\
\cdashline{3-20}[1pt/1pt]
  & LwF    & 64.21 & 11.11 & 31.62 & 56.80 & 12.98 & 33.03
           & 58.93 & 11.36 & 30.98 & 56.32 & 12.85 & 32.32
           & 7.51 & -0.22 & 0.64 & -1.67 & 1.97 & 2.62 \\
\cdashline{3-20}[1pt/1pt]
  & OGD    & 60.37 & \textbf{12.89} & \textbf{33.54} & \textbf{57.99} & \textbf{13.42} & \textbf{33.89}
           & 57.12 & \textbf{12.92} & \textbf{32.74} & \textbf{56.56} & 12.74 & 32.37
           & 3.15 & -0.05 & 0.49 & -1.07 & \textbf{2.30} & \textbf{2.92} \\
\bottomrule
\end{tabular}}
\caption{\textbf{Main Results of CL Strategies on \Benchmark{}, Random vs.\ Semantic Order.}
All reported values of GEC (P/R/F$_{0.5}$), Avg.\ Performance (P/R/F$_{0.5}$), and Backward Transfer (P/R/F$_{0.5}$) are \emph{macro‐averaged} across the 10 disciplines.  
Replay uses a 5\% memory buffer; baselines: SeqFT, LoRA; regularization: EWC, GEM, LwF, OGD.}
\label{tab:main_results}
\end{table*}

\subsection{Experimental Settings}

\paragraph{Base Models.}
We use \texttt{Qwen2.5-7B-Instruct}~\citep{qwen2.5} and \texttt{Llama3-8B-Instruct}~\citep{llama3modelcard} as the backbone models for all experiments. These models were selected for their strong performance on Chinese language tasks and their robust instruction-following capabilities.

\paragraph{Task Sequences and Evaluation.}
To comprehensively assess the impact of task order on continual learning performance, we conducted experiments using two distinct training sequences:
\begin{enumerate}
    \item \textbf{Randomized Order}: The 10 academic disciplines are presented in a randomly shuffled order. To ensure robustness and account for variance, this process is repeated across 5 different random permutations. We report the average results across these runs.
    \item \textbf{Semantically Similar Order}: The tasks are arranged based on their semantic similarity. This sequence simulates a smoother domain transition and is used to investigate the effect of gradual domain shift on catastrophic forgetting.The definition and computation of similarity, as well as the resulting task order, are detailed in the Appendix.

\end{enumerate}
Models are sequentially adapted to the 10 disciplines according to these task sequences. For each run, evaluation is conducted after training on each task, following the procedure outlined in the evaluation section. The final results for the randomized order are averaged over the 5 permutations. Evaluation is performed using official CGEC scoring scripts.

\subsection{Continual Learning Methods}
We investigate the performance of various continual learning methods applied to the domain-specific GEC task. Given that our task requires adapting a large language model to a series of distinct yet related domains, we focus on a strategy combining Parameter-Efficient Tuning (PET) with established continual learning algorithms. To this end, we benchmark the following four categories of adaptation strategies:

\begin{itemize}
     \item \textbf{Sequential Finetuning (SeqFT)}: A naive baseline where the model is trained on each task in sequence without any specific mechanism to retain prior knowledge. This approach provides a lower bound for performance and highlights the problem of catastrophic forgetting.
    \item \textbf{Parameter-Efficient Tuning (LoRA)}: We apply Low-Rank Adaptation with rank 8. This serves as a lightweight adaptation approach.
    \item \textbf{Replay-based Methods}: To mitigate forgetting, we implement experience replay by retaining 2\%, 5\%, or 10\% of training data from previous tasks.
    \item \textbf{Continual Learning Algorithms}: For our single-domain GEC task, we combine Parameter-Efficient Tuning (LoRA) with a set of representative continual learning algorithms to achieve superior results. We evaluate four such approaches: 
    \begin{itemize}
    \item  \textbf{EWC} (Elastic Weight Consolidation)~\citep{Kirkpatrick_Pascanu_Rabinowitz_Veness_Desjardins_Rusu_Milan_Quan_Ramalho_Grabska-Barwinska_et_al._2017}, which regularizes important parameters; 
    \item \textbf{LwF} (Learning without Forgetting)~\citep{Li_Hoiem_2016}, which uses knowledge distillation; 
    \item \textbf{GEM} (Gradient Episodic Memory)~\citep{Lopez-Paz_Ranzato_2017}, which constrains gradient updates; 
    \item \textbf{OGD} (Orthogonal Gradient Descent)~\citep{Farajtabar_Azizan_Mott_Li_2019}, which minimizes task interference through orthogonal updates.
    \end{itemize}
\end{itemize}

\section{Analysis}
\begin{table*}[htb!]
\centering
\setlength{\tabcolsep}{4pt}
\renewcommand{\arraystretch}{1.4}

\resizebox{\textwidth}{!}{%
\begin{tabular}{l l
                *{3}{c}  
                *{3}{c}  
                *{3}{c}  
                *{3}{c}  
                *{3}{c}  
                *{3}{c}  
               }
\toprule
\multirow{2}{*}{\textbf{Model}} & \multirow{2}{*}{\textbf{Buffer}}
  & \multicolumn{3}{c}{\textbf{GEC (Rnd)}} & \multicolumn{3}{c}{\textbf{GEC (Sem)}}
  & \multicolumn{3}{c}{\textbf{AvgPerf (Rnd)}} & \multicolumn{3}{c}{\textbf{AvgPerf (Sem)}}
  & \multicolumn{3}{c}{\textbf{BWT (Rnd)}}    & \multicolumn{3}{c}{\textbf{BWT (Sem)}}    \\
\cmidrule(lr){3-5}\cmidrule(lr){6-8}
\cmidrule(lr){9-11}\cmidrule(lr){12-14}
\cmidrule(lr){15-17}\cmidrule(lr){18-20}
 & 
  & P & R & F$_{0.5}$ & P & R & F$_{0.5}$ 
  & P & R & F$_{0.5}$ & P & R & F$_{0.5}$ 
  & P & R & F$_{0.5}$ & P & R & F$_{0.5}$ \\
\midrule
\multirow{3}{*}{\makecell{\textbf{Qwen2.5}\\\textbf{7B-Instruct}}}
  & 2\,\%   
      & 56.69 & 11.15 & 30.02 & \textbf{50.56} & 12.15 & \textbf{30.12} 
      & 49.56 & 12.27 & 29.25 & 47.94 & \textbf{14.44} & \textbf{31.27 }
      & 5.39 & -1.26 & \textbf{-0.61} & \textbf{-0.54} & -0.68 & -0.73 \\
\cdashline{3-20}[1pt/1pt]
  & 5\,\%   
      & \textbf{58.78} & \textbf{11.49} & \textbf{31.13} & 47.04 & 11.22 & 26.90
      & \textbf{50.85} & \textbf{12.58} & \textbf{29.93} & 48.00 & 13.89 & 30.89
      & \textbf{5.75} & -1.73 & -0.65 & -4.31 & -1.64 & -4.17 \\
\cdashline{3-20}[1pt/1pt]
  & 10\,\%  
      & 53.45 & 11.48 & 29.92 & 47.51 & \textbf{12.18} & 29.13
      & 49.45 & 12.24 & 29.50 & \textbf{48.56} & 13.01 & 30.21 
      & 2.40 & \textbf{-1.13} & -0.64 & -2.97 & \textbf{-0.67} & \textbf{-0.59} \\
\midrule
\multirow{3}{*}{\makecell{\textbf{LLaMA3}\\\textbf{8B-Instruct}}}
  & 2\,\%   
      & \textbf{45.24} & 8.60 & \textbf{23.71} & \textbf{43.47} & 9.33 & \textbf{24.62} 
      & \textbf{33.02} & 10.37 & \textbf{21.68} & \textbf{34.71} & \textbf{11.00} & \textbf{23.50 }
      & \textbf{13.36} & -2.06 & 1.32 & \textbf{9.20} & -1.10 & 1.31 \\
\cdashline{3-20}[1pt/1pt]
  & 5\,\%   
      & 37.31 & \textbf{10.07} & 23.70 & 34.32 & \textbf{9.56} & 22.17
      & 30.00 & \textbf{10.39} & 20.79 & 27.90 & 10.77 & 20.46
      & 7.41 & \textbf{-0.60} & \textbf{2.50} & 7.73 & \textbf{-0.52} & \textbf{2.64} \\
\cdashline{3-20}[1pt/1pt]
  & 10\,\%  
      & 37.25 & 8.85 & 22.06 & 33.54 & 8.4 & 20.62 
      & 29.36 & 9.41 & 19.93 & 30.61 & 10.02 & 21.01 
      & 9.38 & -1.03 & 2.11 & 4.26 & -0.94 & 0.74 \\
\bottomrule
\end{tabular}%
} 
\caption{\textbf{Replay Results on \Benchmark{} Benchmark.}
We report average GEC (P / R / F$_{0.5}$), Avg.\ Performance (P / R / F$_{0.5}$) and Backward Transfer (P / R / F$_{0.5}$)
for three replay buffer sizes (2\%, 5\%, 10\%) under Random vs.\ Semantic task orders.}
\label{tab:replay_results}
\end{table*}

Our analysis of experiments at the \Benchmark{} benchmark reveals several critical insights into the efficacy of continual learning (CL) strategies for large language models that perform Grammatical Error Correction (GEC). We discuss the three dimensions (GEC, AvgPerf, and BWT) and the impact of the two task orders. The results highlight not only the performance of different methods but also the crucial impact of model choice and task ordering.

\subsection{Performance of Foundation Models and Baselines}

The choice of foundation model has a profound effect on performance. Qwen2.5-7B-Instruct consistently and substantially outperforms LLaMA3-8B-Instruct in all metrics (GEC, Average Performance, and Backward Transfer). For example, according to a semantic task order, the best AvgPerf F$_{0.5}$ for Qwen (37.08) is significantly higher than the best of LLaMA (32.38). This suggests that the Qwen model's pretraining on a multilingual corpus, which likely includes extensive Chinese language data, provides a stronger and more transferable inductive bias for the GEC task than the LLaMA model.

Furthermore, the LoRA baseline, a parameter-efficient fine-tuning method, consistently yields a significant performance boost over the naive SeqFT baseline for both models. This establishes LoRA as a strong, non-trivial baseline for this domain. LoRA achieves positive backward transfer (BWT) in a randomized task order (F$_{0.5}$: 4.01) and degrades slightly in a semantic order (F$_{0.5}$: 3.00), but still consistently outperforms SeqFT in this regard.

\subsection{Analysis of Continual Learning Strategies}

Overall, our results indicate that regularization-based continual learning (CL) methods generally outperform baselines such as SeqFT and Replay in mitigating catastrophic forgetting on the GEC task. Although all CL methods are built upon a LoRA-fine-tuned backbone, they further optimize those parameters and consistently achieve higher GEC and AvgPerf scores. In fact, we observe that CL methods account for nearly all optimal entries inC and AvgPerf metrics, highlighting the necessity and effectiveness of continuocontinuousng in this domain.

Regarding the Backward Transfer (BWT) metric, we find that most CL methods exhibit positive BWT values—an improvement over SeqFT and Replay, which suffer from negative BWT. Although the BWT scores of CL methods may not surpass those of LoRA in certain dimensions (e.g., BWT(sem) Recall and F$_{0.5}$), this trade-off reflects a more balanced optimization across multiple training tasks. These methods are designed not merely for retention, but for maintaining overall task stability and reducing the risk of catastrophic forgetting.

\subsubsection{OGD: Prioritizing Overall Performance.} A deeper analysis reveals that OGD is among the top-performing strategies, particularly in terms of average performance. It achieves the highest AvgPerf scores for the Qwen model under both task orders—F$_{0.5}$: 34.97 (randomized) and F$_{0.5}$: 37.08 (semantic). Additionally, it delivers the best GEC(sem) result with a score of 36.53. However, this strong forward learning comes at the cost of lower BWT values, suggesting that OGD focuses more on acquiring new tasks than on preserving older knowledge. This aligns with its core mechanism—enforcing gradient orthogonality—which prevents interference with prior tasks but may limit active reuse of past knowledge.
\subsubsection{GEM: Excelling in Knowledge Retention.} In contrast, GEM demonstrates strong backward transfer performance. It achieves the highest BWT score for the Qwen model under the randomized task order (F$_{0.5}$: 4.11) and shows robust BWT under semantic order as well. These results indicate that GEM effectively preserves knowledge across tasks, especially when they are semantically related. This supports the underlying design of GEM, which constrains gradient updates to protect earlier tasks—making it a strong candidate when long-term retention is a critical objective.

\subsection{Impact of Task Order on Continual Learning}

The ordering of tasks exerts a nuanced and non-uniform impact on continual learning performance, with distinct trends observed across different evaluation dimensions.

For Average Performance (AvgPerf), semantic task order generally leads to improvements in recall and F$_{0.5}$ scores across most continual learning strategies, yet often results in a noticeable decline in precision. For example, in the case of OGD on Qwen, recall improves from 36.13 to 39.17, and F$_{0.5}$ improves from 34.97 to 37.08, while precision drops from 34.90 to 32.36. This pattern is also observable in GEM, EWC, and Replay. One possible explanation is that semantically similar tasks encourage the model to generalize more broadly, leading to increased coverage (recall) at the expense of specificity (precision). This trade-off may reflect greater ambiguity in correction boundaries across similar domains.

In contrast, random task order presents a more diverse distribution of topics, which might promote sharper decision boundaries for each individual task, yielding higher precision but lower generalization across tasks.

Turning to the Backward Transfer (BWT) metric, we observe an intriguing divergence between models. For the Qwen model, BWT often \emph{declines} under semantic order—for instance, EWC's F$_{0.5}$ drops from 4.00 (random) to 2.77 (semantic), and Replay shows a more severe decline from $-$0.65 to $-$4.17. This suggests that Qwen’s representation space may be more sensitive to semantic redundancy, leading to overfitting or interference when similar tasks are presented consecutively.

In contrast, the LLaMA model often shows \emph{improved} BWT under semantic ordering. For example, GEM improves from 2.41 to 3.14, and LoRA improves from 2.81 to 3.56. This indicates that LLaMA may benefit more from task similarity when transferring knowledge backward. One plausible explanation lies in the model's pretraining: Qwen, being heavily pretrained on Chinese corpus, may already exhibit strong task-specific inductive biases that interfere under semantic overlap, while LLaMA, being more general, may leverage this similarity for consolidation.

These results reveal that the effect of task order is not universal, but is deeply intertwined with the properties of the model and strategy used. Semantic ordering is beneficial for average performance and recall, but it may introduce degradation in precision and backward transfer depending on model-specific characteristics.


\subsection{Analysis of Replay Strategy}

We further examine the impact of replay buffer size (2\%, 5\%, 10\%) on continual learning performance in Table~\ref{tab:replay_results}. Surprisingly, increasing the buffer size does not consistently improve performance; instead, we observe a clear performance peak at either 2\% or 5\% buffer for most metrics and models.

\subsubsection{Qwen2.5 Performance Trends.}
For Qwen2.5, the highest GEC F$_{0.5}$ (31.13) is achieved under random order with a 5\% buffer, but performance declines when increasing to 10\% (29.92). A similar trend is observed in AvgPerf and semantic GEC. This suggests that larger replay buffers may introduce noise or redundant samples, weakening the model's focus on current tasks. Additionally, BWT scores remain negative across all settings, with semantic order yielding especially poor results at 5\% buffer (F$_{0.5}$: $-4.17$), indicating that replay alone is insufficient to prevent forgetting for Qwen.

\subsubsection{LLaMA3 Performance Trends.}
In contrast, LLaMA3 benefits more from replay. The best GEC and AvgPerf results appear at 2\% buffer, with stable BWT improvements as the buffer increases. For instance, BWT F$_{0.5}$ improves from 1.31 (2\%) to 2.64 (5\%) under semantic order. This suggests that LLaMA3 is more effective at leveraging limited replay data to retain past knowledge, possibly due to better generalization capabilities.

These findings highlight that simply increasing the replay buffer does not guarantee better performance. In some cases, it may even degrade results due to capacity constraints or replay noise. Moreover, the effectiveness of replay is model-dependent: while Qwen suffers from persistent forgetting, LLaMA exhibits moderate knowledge retention gains. This underscores the necessity of combining replay with more adaptive mechanisms in real-world setups.

\section{Conclusion}

This paper presents \textbf{CL2GEC}, the first continual learning benchmark for Chinese grammatical error correction (CGEC) in academic writing. Unlike prior CGEC datasets that focus on static learner corpora, CL2GEC simulates domain-incremental learning through a multi-disciplinary corpus covering 10 scientific domains. The benchmark supports sequential training and fine-grained evaluation of model forgetting, adaptation, and transfer.
We provide a high-quality dataset of 10,000 human-annotated sentences and define evaluation protocols and continual learning metrics tailored to the GEC task. We benchmark strong language models using parameter-efficient tuning and four representative CL algorithms—EWC, GEM, LwF, and OGD. Our experiments reveal that regularization- and projection-based methods consistently outperform naive sequential fine-tuning and simple replay strategies, though performance varies with task order and model backbone.
We hope CL2GEC lays the groundwork for future research in adaptive GEC systems, encouraging the development of lifelong writing assistants that can generalize across specialized academic domains.




\bibliography{custom}
\section{Reproducibility Checklist}

\paragraph{This paper:}
\begin{itemize}
    \item Includes a conceptual outline and/or pseudocode description of AI methods introduced. \textbf{Yes}.
    \item Clearly delineates statements that are opinions, hypotheses, and speculation from objective facts and results. \textbf{Yes}.
    \item Provides background references or pedagogical pointers for unfamiliar readers. \textbf{Yes}.
\end{itemize}

\paragraph{Does this paper make theoretical contributions?} \textbf{No}.

\paragraph{Does this paper rely on one or more datasets?} \textbf{Yes}.

\begin{itemize}
    \item Motivation is provided for the choice of datasets. \textbf{Yes}.
    \item All novel datasets introduced in this paper are described in the main paper and/or data appendix. \textbf{Yes}.
    \item All novel datasets will be made publicly available upon publication with a license that permits unrestricted research usage. \textbf{Yes}.
    \item All datasets drawn from prior work are cited appropriately. \textbf{Yes}.
    \item All external datasets used are publicly available. \textbf{Yes}.
    \item All datasets that are not publicly available are described in detail, along with rationale for their usage. \textbf{N/A}.
\end{itemize}

\paragraph{Does this paper include computational experiments?} \textbf{Yes}.

\begin{itemize}
    \item Pre-processing scripts will be released alongside the code. \textbf{No (will release after review)}.
    \item All source code used in experiments will be made publicly available upon publication. \textbf{Yes}.
    \item Code is documented with comments and references to corresponding sections in the paper. \textbf{Yes}.
    \item Any stochasticity in training is controlled (e.g., random seed described). \textbf{Yes}.
    \item Hardware and software specifications are reported (e.g., GPU, RAM, CUDA, library versions). \textbf{Partial}.
    \item Evaluation metrics are formally defined and motivated. \textbf{Yes}.
    \item Number of experiment runs and variance across runs is reported. \textbf{Partial}.
    \item Results are primarily summarized using mean performance. \textbf{Yes}.
    \item Significance testing or confidence intervals are provided. \textbf{No}.
    \item Final hyperparameter values used for all models are listed. \textbf{Yes}.
    \item Hyperparameter search ranges and selection criteria are described. \textbf{Yes}.
\end{itemize}

\end{document}